\documentclass{article}
\usepackage{spconf,amsmath,graphicx}
\usepackage{booktabs}
\usepackage{float}
 \usepackage{multirow} 
\usepackage{bm}

\title{MeniMV: A Multi-view Benchmark for Meniscus Injury Severity Grading}

\name{Shurui Xu$^{1}$, Siqi Yang$^{2}$, Jiapin Ren$^{3}$, Zhong Cao$^{4}$, 
Hongwei Yang$^{5}$, Mengzhen Fan$^{6}$, Yuyu Sun$^{5}$, Shuyan Li$^{1}$}

\address{$^{1}$ Electronics, Electrical Engineering and Computer Science, Queen's University Belfast\\
$^{2}$ Department of Radiology, Affiliated Hospital 2 of Nantong University\\
$^{3}$ Xuhai College, China University of Mining and Technology\\
$^{4}$ Heidelberg Institute of Global Health, Faculty of Medicine and University Hospital, Heidelberg University\\
$^{5}$ Department of Orthopaedics, Affiliated Nantong Hospital 3 of Nantong University\\
$^{6}$ HSBC Business School, Peking University}

\begin{document}
%
\maketitle
\begin{abstract}
Precise grading of meniscal horn tears is critical in knee injury diagnosis but remains underexplored in automated MRI analysis. Existing methods often rely on coarse study-level labels or binary classification, lacking localization and severity information. In this paper, we introduce MeniMV, a multi-view benchmark dataset specifically designed for horn-specific meniscus injury grading. MeniMV comprises 3,000 annotated knee MRI exams from 750 patients across three medical centers, providing 6,000 co-registered sagittal and coronal images. Each exam is meticulously annotated with four-tier (grade 0–3) severity labels for both anterior and posterior meniscal horns, verified by chief orthopedic physicians. Notably, MeniMV offers more than double the pathology-labeled data volume of prior datasets while uniquely capturing the dual-view diagnostic context essential in clinical practice. To demonstrate the utility of MeniMV, we benchmark multiple state-of-the-art CNN and Transformer-based models. Our extensive experiments establish strong baselines and highlight challenges in severity grading, providing a valuable foundation for future research in automated musculoskeletal imaging.
\end{abstract}
\begin{keywords}
Meniscus Injury Grading, Knee MRI, Deep Learning, Medical Image Analysis, Dataset.
\end{keywords}
\section{Introduction}

\begin{figure}[h]
\centering
\includegraphics[width=0.5\textwidth]{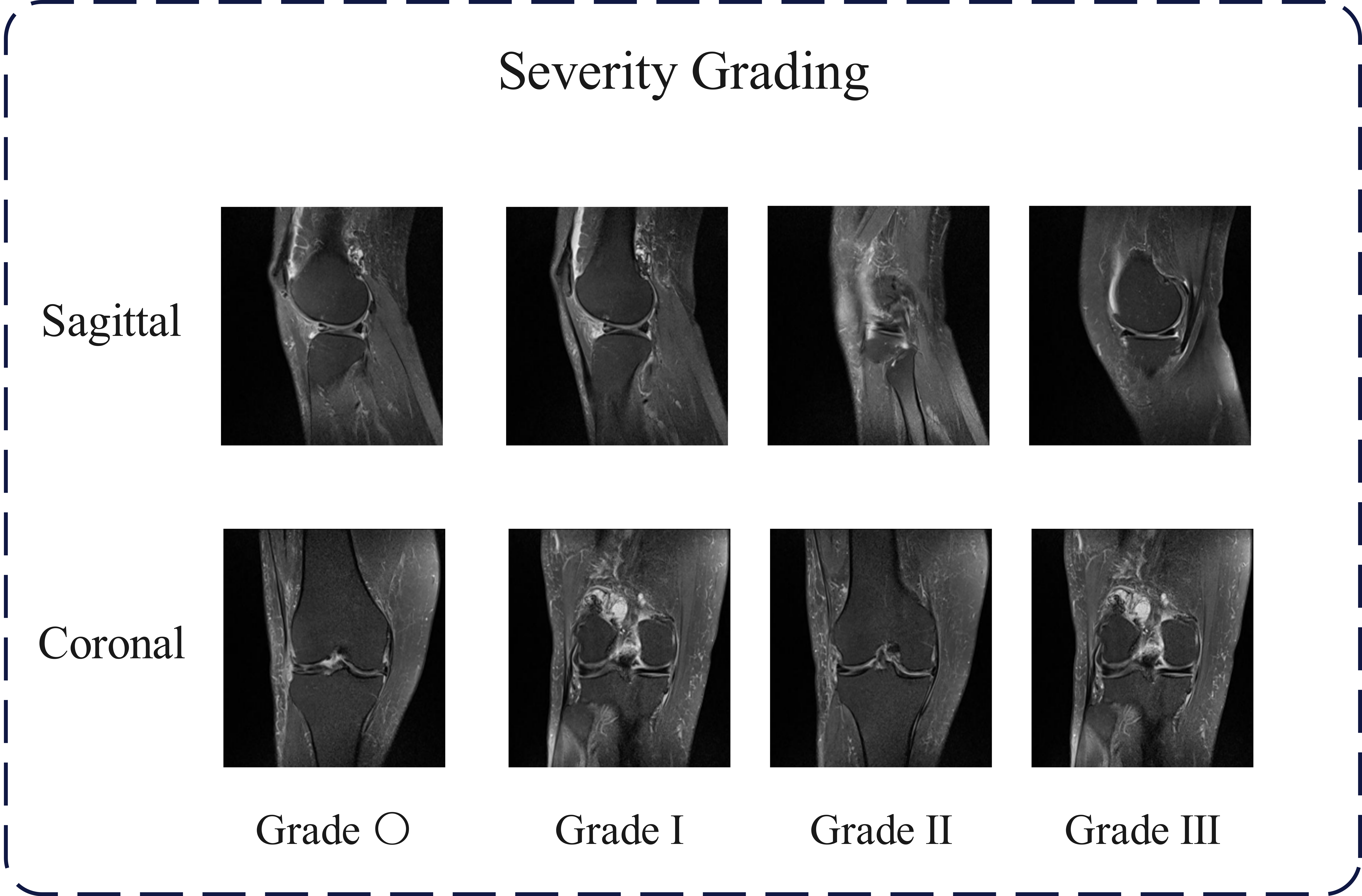}
\caption{Example from the MeniMV dataset. Each case includes two views—sagittal and coronal—with anterior and posterior meniscal horns annotated across four injury grades.}
\label{fg1}
\end{figure}

The knee meniscus plays a vital role in knee function by redistributing force, absorbing impact and stabilising movement between the tibia and femur, thus helping to protect the articular cartilage from early degeneration. Tears of the anterior or posterior horn of the meniscus are one of the most common reasons for sports medicine referrals, accounting for one-third of knee arthroscopy procedures each year. Magnetic Resonance Imaging (MRI) is currently considered the gold standard for diagnosing knee injuries due to its excellent soft tissue contrast and ability to capture images from multiple angles. However, interpreting a full set of MRIs requires significant expertise and time, which can lead to inconsistencies between different interpreters, increased costs, and delays in treatment decisions. Automated systems powered by deep learning have the potential to provide more consistent and objective assessments, allowing for more accurate treatment.

Deep learning has demonstrated significant potential in automating the diagnosis of knee injuries. Early work primarily focused on cartilage segmentation and lesion detection using convolutional neural networks (CNNs)~\cite{R1, R2, R3}. A major advancement came with the introduction of MRNet~\cite{R4}, which utilized multi-view MRI scans (sagittal, coronal, axial) to classify entire knee exams as either "normal" or "abnormal." While this demonstrated the feasibility of deep learning for knee assessment, MRNet was limited by its reliance on coarse, study-level labels (e.g., "abnormal," "ACL tear," "meniscal tear"), lacking finer-grained information such as tear location or severity. Subsequent efforts like the fastMRI~\cite{R5} addressed data scarcity by releasing large volumes of raw k-space data and images. This effort was later extended by fastMRI+~\cite{R6}, which included pathology labels. However, annotations are restricted to the coronal plane, crucially treating meniscal tears as a binary classification problem (i.e. presence or absence), overlooking the important aspect of injury grading.

To address these limitations, we present MeniMV, a multi-view benchmark dataset specifically designed for meniscus injury grading. The dataset comprises MRI scans from 750 patients collected across three medical centers, with all images independently validated by a panel of expert clinicians and radiologists to ensure high-quality and reliable annotations. Each examination includes horn-specific, four-tier tear labels (grades 0–3), capturing the severity of damage in the anterior or posterior meniscal horn. As a result, MeniMV contains 3,000 annotated knee examinations, each with paired sagittal and coronal T2-weighted MRI series, totaling 6,000 co-registered scans. As illustrated in Figure~\ref{fg1}, MeniMV provides annotations for both sagittal and coronal views, covering the anterior and posterior meniscal horns across four injury grades. Notably, MeniMV not only more than doubles the volume of pathology-labeled data available in MRNet and fastMRI combined, but also uniquely offers the clinically essential dual-view context routinely used by radiologists during diagnosis. Leveraging this comprehensive dataset, we formulate meniscal injury assessment as a dual-head classification task. We further perform extensive experiments using standard classification backbones to evaluate their effectiveness in fine-grained severity grading on the MeniMV benchmark.

Our main contributions are summarized as follows:
\begin{itemize}
\item We released MeniMV, a multi-view knee MRI dataset with sagittal and coronal images, annotated for anterior/posterior horn tears across four severity levels (0–3), offering a high-quality benchmark for the research community. As far as we know, MeniMV is the first resource to provide horn-specific severity grades on paired sagittal-coronal planes.

\item We benchmarked several popular backbone architectures for automated injury severity grading on the proposed dataset, including both CNN-based and Transformer-based frameworks.

\end{itemize}

\section{MeniMV Dataset}

\subsection{Data Collection and Labeling}
We retrospectively curated 750 knee MRI studies from three hospitals (405 females, 345 males; mean age $55.6 \pm 12.7$ years; range 14–82) under a unified protocol. To enhance sensitivity to intrameniscal signal change, only fat-suppressed T2-weighted (T2WI-FS) sequences were included (reported sensitivity $\sim$94\% for meniscal lesions~\cite{R7}). All DICOMs were de-identified with a standardized pipeline (DicomAnonymizer) and metadata archived for reproducibility~\cite{R8, R9}.
\begin{figure}[h]
\centering
\includegraphics[width=0.5\textwidth]{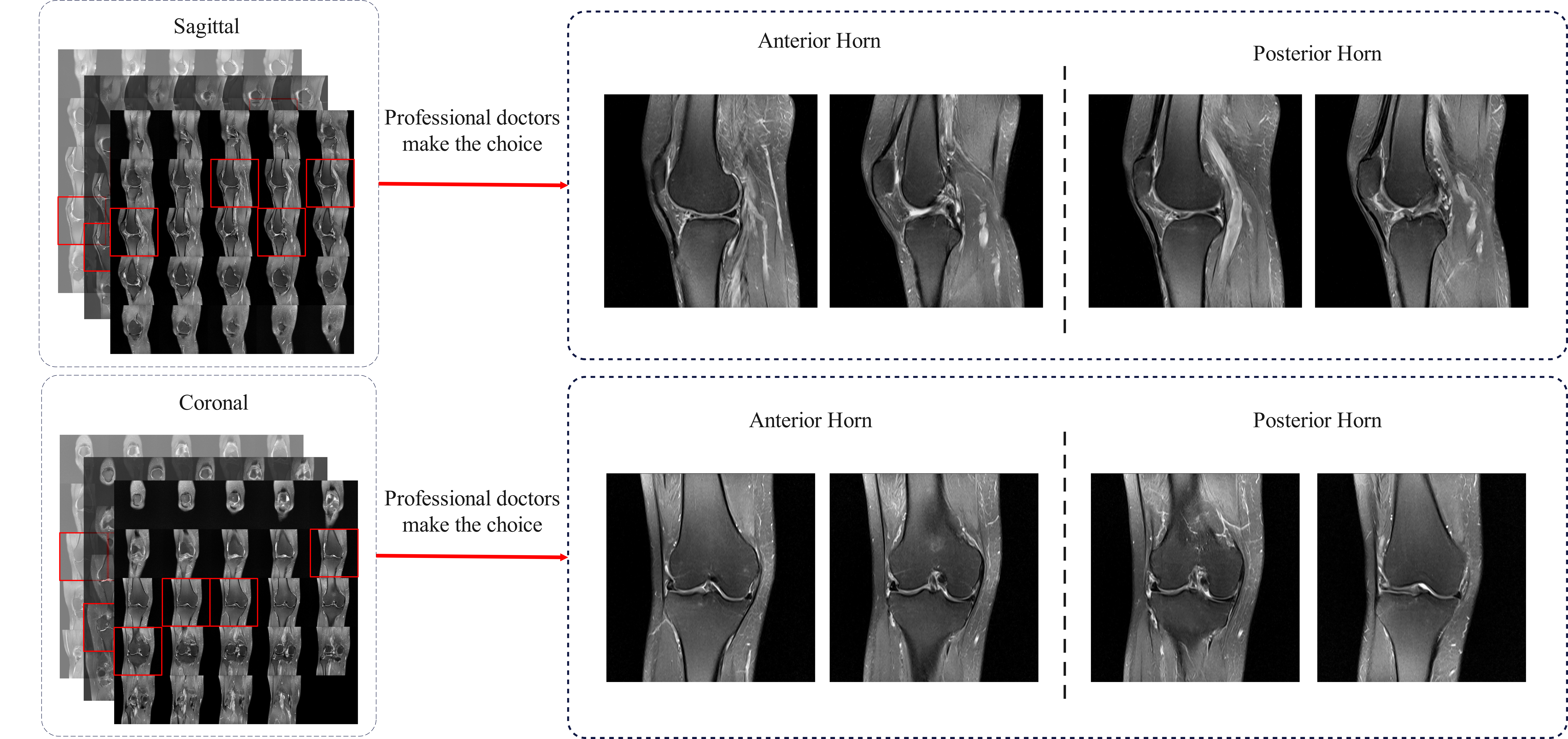}
\caption{The selection procedure of MeniMV. The four most diagnostically relevant slices from both sagittal and coronal MRI series are selected by chief orthopedic physicians.}
\label{fg1}
\end{figure}
A panel of six senior orthopedic clinicians (each $\geq$10 years’ experience) independently screened each study focusing on the anterior and posterior horns. For each horn, the most informative sagittal–coronal slice pair was retained; motion-degraded images were excluded. Severity was assigned per image using the Stoller 0--III (0–3) scale with double validation. In total, MeniMV contains 3{,}000 expert-annotated images (1{,}500 sagittal–coronal pairs; two pairs per patient).

\subsection{Dataset Properties and Context}
MeniMV provides four images per patient (two sagittal and two coronal; one pair per horn), enabling consistent multi-view learning under a single labeling scheme. Across the cohort (age 14–82; mean $55.6 \pm 12.7$), Grade 0 cases are more frequent at younger ages and decline with age; Grades 1–2 are relatively uniform; Grade 3 rises notably after 50 and concentrates in the 60–70 range, consistent with age-related degeneration.

The grade distribution is: 1{,}331 Grade 0 (44.4\%), 502 Grade 1 (16.7\%), 306 Grade 2 (10.2\%), and 861 Grade 3 (28.7\%). The scarcity of Grades 1–2 motivates class-balanced training objectives and ordinal-aware evaluation (e.g., quadratic weighted Cohen’s $\kappa$, grade-wise MAE).

Compared with prior knee MRI resources, MRNet~\cite{R4} aggregates 1{,}370 single-center studies with mixed sequences and single-reader binary labels, and fastMRI targets reconstruction without diagnostic labels. In contrast, MeniMV is multi-center (three hospitals), standardized to T2WI-FS, and supplies.

\section{Multi-view Meniscus Injury Grading}

\subsection{Task Definition}

 We address meniscal injury severity classification from paired MRI views (sagittal and coronal), assigning each case to one of four grades: 0 (normal), 1 (mild), 2 (moderate), or 3 (severe), with position-specific categorization for anterior and posterior horns. Given a pair of views $\bm{x}=\{x_{\text{sagittal}},\,x_{\text{coronal}}\}$ and a slice-level anatomical indicator $p\in\{0,1\}$ (anterior $=1$, posterior $=0$), the objective is to predict an injury grade $y\in\mathcal{Y}=\{0,1,2,3\}$ at the specified position. We formalize this as a mapping $f_{\theta}:(\bm{x},p)\mapsto\mathcal{Y}$, where $\theta$ denotes the model parameters comprising a shared backbone feature extractor and two position-specific classification heads. The inclusion of $p$ conditions the prediction on the target horn, enabling the model to treat anterior and posterior regions independently in light of their distinct clinical failure patterns.

\subsection{Multi-view Encoder}

We utilize a range of architectures as backbone feature extractors, including ResNet-50 \cite{R10,R11}, DenseNet-121 ~\cite{R12,R13,R14}, EfficientNet-B0 \cite{R15,R16,R17}, and \textbf{Vision Transformer (ViT-B/16)}~\cite{R18,R19}. We train our model from scratch using the proposed dataset. Subsequently, we extract feature embeddings from both the sagittal and coronal views, denoted as $\epsilon_{\text{sagittal}}$ and $\epsilon_{\text{coronal}}$, respectively.

While there are multiple ways to fuse these features, we empirically found that concatenation yields the best performance. Then we get the fusion feature as below:

\begin{equation}
   \mathbf{f}_{\text{fusion}} = \left[ \mathbf{\epsilon}_{\text{sagittal}} \parallel \mathbf{\epsilon}_{\text{coronal}} \right], 
\end{equation}

where \( \parallel \) represents concatenation along the feature dimension. This fusion enables the model to leverage complementary anatomical details from both the sagittal and coronal planes, ensuring a more comprehensive understanding of the meniscal injury.

\subsection{Classification Head}

After feature fusion, the concatenated feature vector \( \mathbf{f}_{\text{fusion}} \) is forwarded through a classification head to predict the injury severity. We introduce a dual-classification head for grading in the anterior and posterior regions of the meniscus separately. Specifically, the model uses two separate fully connected layers:

\begin{equation}
P =
\begin{cases}
\mathrm{softmax}\left(W_a \, \mathbf{f}_{\text{fusion}} + b_a\right), & \text{if } p = 1, \\
\mathrm{softmax}\left(W_p \, \mathbf{f}_{\text{fusion}} + b_p\right), & \text{if } p = 0,
\end{cases}
\end{equation}
where \(W_a, W_p\) and \(b_a, b_p\) are the parameters for the anterior and posterior classifiers, respectively. This ensures that the model learns region-specific decision boundaries.

\subsection{Loss Function}

We optimize a hybrid objective that couples class-imbalance–aware Focal Loss with a cross-view feature-alignment term. Given a minibatch of $N$ labeled samples, the multi-class Focal Loss is
\begin{equation}
L_{\text{focal}}=-\sum_{i=1}^{N}\lambda_{y_i}\,\big(1-P_{i,y_i}\big)^{\gamma}\,\log\!\big(P_{i,y_i}\big),
\end{equation}
where $P_{i,y}=\mathrm{softmax}(\mathbf{z}_i)_y$ is the predicted probability for class $y$ from logits $\mathbf{z}_i$, $\lambda_{y}\!\ge\!0$ are class-specific weights to counter class imbalance, and $\gamma\!\ge\!0$ controls the down-weighting of well-classified examples (larger $\gamma$ emphasizes harder cases) . To promote coherence between sagittal and coronal representations, we further penalize discrepancies between their latent features via an MSE term:
\begin{equation}
L_{\text{MSE}}=\frac{1}{N}\sum_{i=1}^{N}\left\|\mathbf{f}^{(i)}_{\text{sagittal}}-\mathbf{f}^{(i)}_{\text{coronal}}\right\|_2^{2}.
\end{equation}
The overall training objective is a weighted sum of the two components,
\begin{equation}
L_{\text{total}}=\alpha\,L_{\text{focal}}+\beta\,L_{\text{MSE}},
\end{equation}
with $\alpha,\beta\!\ge\!0$ controlling the trade-off between label-distribution sensitivity and cross-view alignment.

\section{Experimental Analysis}

\subsection{Backbone and Method Analysis}
Table~\ref{tab:backbones_extended} contrasts generic CNNs, domain-specific methods, and modern Transformers. Among generic CNNs trained from scratch, DenseNet-121 yields the strongest results (67.83\% Acc). However, domain-specific methods (MRNet~\cite{R4}, 2.5D ResNet Fusion~\cite{R11}, and DeepKnee~\cite{R3}) significantly outperform these generic baselines; specifically, DeepKnee achieves 74.22\% Acc, surpassing DenseNet-121 by a notable margin (+6.39\%) due to its specialized ROI localization and fusion design. Nevertheless, pretrained Transformers demonstrate the superior discrimination: the Swin-UNETR encoder attains the state-of-the-art performance (76.92\% Acc, 0.6790 Macro-F1), outperforming even the specialized DeepKnee method. This progression suggests that while domain-specific engineering is effective, the large-scale representation learning capabilities of modern Transformers provide the most robust foundation for meniscal injury grading.

\begin{table}[htbp]
\centering
\vspace{-0.3cm}
\caption{Performance comparison of generic backbones, domain-specific methods, and modern pretrained architectures.}
\label{tab:backbones_extended}
\begin{tabular*}{\linewidth}{@{\extracolsep{\fill}}lccc}
\toprule
Method / Backbone & Acc (\%) & Macro-F1 & MAE \\
\midrule
\multicolumn{4}{l}{\textit{Generic CNNs (Scratch)}} \\
ResNet-50 & 57.67 & 0.4667 & 0.91 \\
ResNeXt-50 (32$\times$4d) & 59.92 & 0.4874 & 0.88 \\
EfficientNet-B0 & 55.00 & 0.4362 & 0.95 \\
ShuffleNet-v2 & 54.83 & 0.3949 & 0.99 \\
MobileNet-v2 & 53.89 & 0.3825 & 1.01 \\
ConvNeXt-T & 63.78 & 0.5311 & 0.81 \\
DenseNet-121 & 67.83 & 0.5730 & 0.74 \\
ViT-B/16 (scratch) & 50.33 & 0.2907 & 1.10 \\
\midrule
\multicolumn{4}{l}{\textit{Domain-Specific Methods}} \\
MRNet (Adapted) & 69.45 & 0.5912 & 0.68 \\
2.5D ResNet Fusion & 71.80 & 0.6150 & 0.63 \\
DeepKnee (Adapted) & 74.22 & 0.6385 & 0.58 \\
\midrule
\multicolumn{4}{l}{\textit{Modern Pretrained Architectures}} \\ %
ConvNeXt-V2-B (pt) & 74.85 & 0.6550 & 0.55 \\
Swin-T (pt) & 75.42 & 0.6610 & 0.54 \\
Swin-B (pt) & 76.38 & 0.6730 & 0.52 \\
ViT-B/16 (MAE/DINO pt) & 73.11 & 0.6400 & 0.57 \\
Swin-UNETR encoder (pt) & \textbf{76.92} & \textbf{0.6790} & \textbf{0.51} \\
\bottomrule
\end{tabular*}
\vspace{-0.3cm}
\end{table}

\subsection{Fusion Strategy Analysis}
To assess the effect of feature fusion in our dual-view framework, we evaluate three canonical strategies applied after independently encoding the sagittal and coronal views with a shared DenseNet-121 backbone: additive fusion, which aggregates the two feature vectors via element-wise summation; concatenation, which joins the representations channel-wise; and attention-based fusion, which introduces a gated self-attention module to adaptively reweight and combine the sagittal and coronal features.

As shown in Table~\ref{tab:fusion_strategy}, the concatenation strategy outperforms both additive and attention-based fusion. Additive fusion offers simplicity but assumes equal importance and alignment between the sagittal and coronal features, which may not hold given their distinct anatomical information. Attention-based fusion improves over simple addition by learning dynamic weighting across modalities, but introduces extra parameters and computation, with only modest performance gains.
\begin{table}[h]
\centering
\vspace{-0.3cm}
\caption{Comparison of feature fusion strategies.}
\label{tab:fusion_strategy}
\begin{tabular}{lcc}
\toprule
\textbf{Fusion Strategy} & \textbf{Accuracy (\%)} & \textbf{Macro-F1} \\
\midrule
Additive Fusion & 70.02 & 0.6021 \\
Attention-based Fusion & 72.46 & 0.6228 \\
Concatenation & \textbf{73.63} & \textbf{0.6347} \\
\bottomrule
\end{tabular}
\vspace{-0.3cm}
\end{table}

\subsection{Cross-center Robustness (LOCO)}
To assess out-of-distribution robustness, we adopted a leave-one-center-out (LOCO) protocol. As shown in Table~\ref{tab:loco}, domain-specific methods generalize better than standard CNNs; specifically, DeepKnee achieves an average Macro-F1 of 0.625 and Acc of 72.5\%, demonstrating that incorporating anatomical priors helps mitigate scanner variations. However, modern pretrained Transformers generalize substantially better than both generic and domain-specific baselines. Swin-UNETR achieves the highest average Macro-F1 of 0.641 and reduces the MAE to 0.56, with lower variance across centers. This indicates that while domain-specific design improves robustness, the self-attention mechanism combined with large-scale pretraining offers the most effective defense against cross-center protocol shifts.

\begin{table}[h]
\centering
\vspace{-0.3cm}
\caption{LOCO performance comparison including domain-specific baselines. Macro-F1 per target center; Acc/MAE averaged across centers.}
\label{tab:loco}
\setlength{\tabcolsep}{3pt} 
\resizebox{\columnwidth}{!}{%
\begin{tabular}{lcccccc}
\toprule
Model & MF1-A & MF1-B & MF1-C & MF1-Avg & Acc-Avg (\%) & MAE-Avg \\
\midrule
DenseNet-121            & 0.551 & 0.563 & 0.532 & 0.549 & 65.4 & 0.79 \\
ConvNeXt-T              & 0.574 & 0.586 & 0.557 & 0.572 & 61.9 & 0.83 \\
\midrule
MRNet (Adapted)         & 0.578 & 0.589 & 0.558 & 0.575 & 68.5 & 0.69 \\
2.5D ResNet Fusion      & 0.601 & 0.612 & 0.581 & 0.598 & 70.2 & 0.65 \\
DeepKnee (Adapted)      & 0.628 & 0.639 & 0.608 & 0.625 & 72.5 & 0.60 \\
\midrule
Swin-T (pt)             & 0.620 & 0.631 & 0.602 & 0.618 & 72.1 & 0.59 \\
Swin-B (pt)             & 0.636 & 0.644 & 0.615 & 0.632 & 73.0 & 0.57 \\
Swin-UNETR encoder (pt) & \textbf{0.645} & \textbf{0.653} & \textbf{0.624} & \textbf{0.641} & \textbf{73.6} & \textbf{0.56} \\
\bottomrule
\end{tabular}%
}
\vspace{-0.3cm}
\end{table}

\subsection{Demographic Robustness}

Finally, we examine sex- and age-stratified results for the best model (Table~\ref{tab:fairness}). Gaps are small but non-negligible: male vs.\ female Macro-F1 differs by 0.018 with a 0.04 MAE gap; across age bands, the Macro-F1 gap is 0.025 and MAE gap is 0.06. These patterns are consistent with an age-dependent prevalence of severe degeneration and suggest that mild post-hoc calibration or reweighting could further narrow disparities without sacrificing overall accuracy.
\begin{table}[htbp]
\centering
\vspace{-0.3cm}
\caption{Demographic robustness for the best model (Swin-UNETR encoder).}
\label{tab:fairness}
\begin{tabular}{lcccc}
\toprule
Subgroup & Macro-F1 & MAE & $\Delta$F1 & $\Delta$MAE \\
\midrule
Male & 0.648 & 0.53 & \multirow{2}{*}{0.018} & \multirow{2}{*}{0.04} \\
Female & 0.666 & 0.49 & & \\
\midrule
$Age < 40$
 & 0.642 & 0.56 & \multirow{3}{*}{0.025} & \multirow{3}{*}{0.06} \\
Age 40–59 & 0.667 & 0.50 & & \\
Age $\ge$ 60 & 0.655 & 0.52 & & \\
\bottomrule
\end{tabular}
\vspace{-0.3cm}
\end{table}

\section{Conclusion}

Our work introduces MeniMV, a large-scale, dual-view knee MRI dataset featuring horn-specific, four-level severity annotations, curated by experienced musculoskeletal radiologists. We benchmark several widely used backbone architectures—including both CNN-based and Transformer-based models—for automated injury grading on this dataset. Looking ahead, we plan to explore cross-modal integration with clinical metadata and extend our framework to other joint structures.

\clearpage
\bibliographystyle{IEEEbib}
\bibliography{refs}

\end{document}